\documentclass[conference]{IEEEtran}
\newcommand{\model}{T$^2$-Net}
\newcommand{\tb}{\textbf}
\newcommand{\ti}{\textit}

\newcommand{\bs}{\Psi}
\newcommand{\bsy}{\boldsymbol}

\newcommand{\mc}{\mathcal}

\IEEEoverridecommandlockouts
\usepackage{cite}
\usepackage{amsmath,amssymb,amsfonts}
\usepackage{graphicx}
\usepackage{textcomp}
\usepackage{xcolor}
\usepackage{cancel}
\usepackage{caption}
\usepackage{subfloat}
\usepackage{multirow}
\usepackage{algorithm}
\usepackage{algorithmic}
\usepackage{tabularx}
\usepackage{array}
\usepackage{enumitem}
\usepackage{float}
\usepackage[numbers]{natbib}
\let\sss= \scriptscriptstyle
\usepackage{relsize}
\usepackage{subfigure}
\usepackage{tikz,pgfplots}
\def\BibTeX{{\rm B\kern-.05em{\sc i\kern-.025em b}\kern-.08em
    T\kern-.1667em\lower.7ex\hbox{E}\kern-.125emX}}
\begin{document}

\title{T$^2$-Net: A Semi-supervised Deep Model for \\Turbulence Forecasting}
\author{Denghui Zhang$^1$, Yanchi Liu$^{2*}$\thanks{$^*$Corresponding author.}, Wei Cheng$^2$, Bo Zong$^2$, Jingchao Ni$^2$, \\
Zhengzhang Chen$^2$, Haifeng Chen$^2$, Hui Xiong$^1$\\
$^1$Rutgers University, USA, \{denghui.zhang, hxiong\}@rutgers.edu\\
$^2$NEC Labs America, USA, \{yanchi, weicheng, bzong, jni, zchen, haifeng\}@nec-labs.com\\
}

\maketitle

\begin{abstract}
Accurate air turbulence forecasting can help airlines avoid hazardous turbulence, guide the routes that keep passengers safe, maximize efficiency, and reduce costs. 
Traditional turbulence forecasting approaches heavily rely on painstakingly customized turbulence indexes, which are less effective in dynamic and complex weather conditions. 
The recent availability of high-resolution weather data and turbulence records allows more accurate forecasting of the turbulence in a data-driven way. 
However, it is a non-trivial task for developing a machine learning based turbulence forecasting system due to two challenges: (1) {Complex spatio-temporal correlations}, turbulence is caused by air movement with complex spatio-temporal patterns, (2) {Label scarcity}, very limited turbulence labels can be obtained. 
To this end, in this paper, we develop a unified semi-supervised framework, \model{}, to address the above challenges. 
Specifically, we first build an encoder-decoder paradigm based on the convolutional LSTM to model the spatio-temporal correlations. 
Then, to tackle the label scarcity problem, we propose a novel {Dual Label Guessing} method to take advantage of massive unlabeled turbulence data. 
It integrates complementary signals from the main \ti{\tb{T}}urbulence Forecasting task and the auxiliary \ti{\tb{T}}urbulence Detection task to generate pseudo-labels,  which are dynamically utilized as additional training data. 
Finally, extensive experimental results on a real-world turbulence dataset validate the superiority of our method on turbulence forecasting.
\end{abstract}


\section{Introduction}
Turbulence is the leading cause of injuries to airline passengers and causes huge loss for airline companies.
According to the U.S. Federal Aviation Administration (FAA), from 1980 through 2008, U.S. air carriers had 234 turbulence accidents, resulting in 298 serious injuries and three fatalities. More lately, since 2009, more than 340 turbulence related injuries have been reported\footnote{https://www.faa.gov/news/fact\_sheets}.
The consequent economic losses are enormous. A vice president of one major air carrier once estimated that it pays out ``tens of millions per year'' for customer injuries, and loses about 7000 days in employee injury-related disabilities \cite{sharman2006integrated}. 
If turbulence can be forecasted accurately so that airlines can reroute ahead, then injuries and property damage can be averted, even lives can be saved.

Despite the benefits, turbulence has been difficult to forecast for being a ``microscale'' phenomenon. 
In the atmosphere, turbulent ``eddies'' vary in size, from hundreds of kilometers down to centimeters. But aircraft bumpiness is most pronounced when the turbulent eddies and aircraft are similar in size. 
It is impossible to directly forecast atmospheric motion at this scale, now or even in the foreseeable future \cite{sharman2006integrated}.
Fortunately, most of the energy associated with turbulent eddies on this scale cascade down from the larger scales of atmospheric motion \cite{dutton1970clear, tung2003k, koshyk2001horizontal}, and these larger scales may be resolved by Numerical Weather Prediction (NWP) models. 
Based on NWP, a variety of \ti{turbulence indexes}, derived from basic weather features, are proposed by meteorologists to estimate the probability of turbulence occurrence \cite{sharman2006integrated}.
While most existing turbulence forecasting methods rely on turbulence indexes, however, we observe that solely using manually crafted features is usually suboptimal, yielding unsatisfactory accuracy. 
Moreover, turbulence indexes have poor generalization power to adapt to new data, especially can not handle more complex situations such as climate change.

On the other hand, prior work has shown that a related problem, i.e., weather forecasting can be solved in a more effective and automatic way leveraging deep learning \cite{hsieh2015inferring, xingjian2015convolutional}, 
whereas, researches on applying advanced machine learning models to turbulence forecasting still remain few.
To this end, we make the very first attempt to leverage deep learning for turbulence forecasting, using turbulence events recorded by pilot reports as ground truth labels.
Nevertheless, we find two inevitable challenges impeding us from building an effective turbulence forecasting system:


\begin{itemize}[leftmargin=*]
	\item \noindent\tb{Complex spatio-temporal correlations}. 
	Turbulence is in nature a \ti{spatio-temporal} phenomenon of air movements. It may occur as a result of various conditions, such as proximity to the jet stream or mountain waves.
	These conditions can actually be captured by certain combinations of meteorological features of the surrounding area and adjacent time slots. Most existing approaches only consider the static features of the target area but neglect the spatio-temporal features of surrounding areas.
	\item \tb{Label scarcity.} Under the paradigm of supervised learning, a large number of turbulence labels are needed to provide signals for training a statistical forecasting model. 
	However, the turbulence label is very \ti{scarce} in the real-world because: (\romannumeral1) turbulence is a rare and anomaly event, (\romannumeral2) it can only be recorded when there is a pilot happens to pass by
	Data with scarce labels, largely limits the power of machine learning.

\end{itemize}

To address the above challenges, we present a unified semi-supervised deep learning framework for turbulence forecasting, namely, \model. 
\model~consists of two modules, i.e., a turbulence forecasting model and a turbulence detection model, which are co-trained in a semi-supervised manner.
The forecasting model is built upon ConvLSTM to learn the complex spatio-temporal patterns for causing turbulence automatically.
To take advantage of massive unlabeled data and alleviate the label scarcity issue, 
we propose a novel {Dual Label Guessing} (DLG) method for data augmentation.
In DLG, 
we introduce an auxiliary task, \tb{T}urbulence Detection, and employ 3D-CNN for this task.
\model~integrates complementary signals from the two tasks to generate more robust pseudo-labels,
which are then utilized as additional data for better generalization ability. 
Finally, we carry out extensive experiments on a real-world dataset to evaluate our model. Results show that \model{} outperforms strong baseline methods in terms of all evaluation metrics (Accuracy, Weighted-Precision, Weighted-Recall, Weighted-F1). Hence the proposed approach can greatly alleviate the problem of spatio-temporal correlation modeling and label scarcity on turbulence forecasting.


\section{Preliminary}
In this section, we introduce several essential preliminaries. First, we give the problem formulation of turbulence forecasting as well as the auxiliary task, turbulence detection. Then we elaborate the features and labels used in our framework. 
\subsection{Turbulence Forecasting}
We formulate the turbulence forecasting problem as a \ti{sequence to sequence multi-class classification} problem. 
That is, given the historical feature cubes (each cube representing a grid-based 3D region) at previous time slots, $\bsy{X_1,X_2,...,X_n}\in \mathbb{R}^{\mc{L\times W\times H\times C}}$, it aims to predict the turbulence levels of all grids in this 3D region at next few time slots, i.e.,  $\bsy{Y_{n+1},Y_{n+2},...,Y_{n+p}}\in \mathbb{R}^{\mc{L\times W\times H\times}4}$. 
$\rm \mc{L\times W\times H}$ indicates the size (number of grids) of the 3D region, $C$ is the number of channels/features per grid, and $4$ denotes the number of turbulence classes. 
Each time slot could be, for example, an hour, 3 hours, or a day.
Let $\bsy{X}=[\bsy{X_1,X_2,...,X_n}]$, $\bsy{Y}=[\bsy{Y_{n+1},Y_{n+2},...,Y_{n+p}}]$, 
we aim to train a statistical model $\mc{F}(\cdot;\theta_{TFN})$, that, given $\bsy{X}$, yields a forecast sequence $\bsy{{P}_{\sss TFN}}$ fitting $\bsy{Y}$:
\begin{equation}
\bsy{{P}_{\sss TFN}}=\mc{F}(\bsy{X};\theta_{\sss TFN})
\end{equation}
In this paper, we set $\rm \mc{L\times W\times H}=10\times 10\times 5$ for computation efficiency and flexibility\footnote{The receptive field of $10\times 10\times 5$ is large enough since each grid has the size of 13km, and turbulence ``eddies'' are normally smaller than 100km \cite{sharman2006integrated}.}. We choose one hour as the length of a time slot,
in other words, we use the previous $n$ hours' feature cubes to forecast the hourly turbulence level of next $p$ hours. 

\subsection{Turbulence Detection}
Turbulence detection is a similar task to forecasting which serves as an auxiliary task in \model. 
Given the NWP forecasted feature cube of a time slot $i$, i.e., $\bsy{X_i}\in \mathbb{R}^{\mc{L\times W\times H\times C}}$, 
turbulence detection aims to predict turbulence conditions of all grids in this 3D region at the {same time slot}, i.e., $\bsy{Y_i} \in \mathbb{R}^{\mc{L\times W\times H\times}4}$. 
In this task,
we aim to train a statistical model $\mc{F}(\cdot;\theta_{TDN})$, that, given $\bsy{X_i}$, return detection result $\bsy{{P}_{i,\sss TDN}}$ fitting $\bsy{Y_i}$:
\begin{equation}
\bsy{P}_{i,\sss TDN}=\mc{F}(\bsy{X_i};\theta_{\sss TDN})
\end{equation}
The detection task differs from forecasting task in two ways: (1) \ti{Synchroneity}, i.e., its features are forecasted based on NWP models and synchronized with the turbulence labels. 
It aims to detect future turbulence using future features while forecasting aims to predict future turbulence using past features.
(2) \ti{Static}, it is also easier since it only predicts one step at one time. 
These two tasks share the same target but have different input features and hold different properties. 
We utilize both turbulence forecasting and detection to provide complementary guidance for the pseudo-label generation. 
\begin{table}
\small
\caption{Raw features and turbulence indexes}
	\centering
	\begin{tabular}{|c|c|c|}
		\hline
		Notation&Name&Unit\\
		\hline
		\hline
		$v_{_U}$&U component of wind & $\rm ms^{-1}$\\
		\hline
		$v_{_V}$&V component of wind & $\rm ms^{-1}$\\
		\hline
		$T$&Temperature& $\rm K$\\
		\hline
		$H$&Relative humidity & \% \\
		\hline
		$V$&Vertical velocity& $\rm Pas^{-1}$\\
		\hline
		$P$&Pressure & $\rm Pa$\\
		\hline
		\hline
		$Ri$&Richardson Number&- \\
		\hline
		$CP$&Colson Panofsky Index&$\rm kt^2$ \\
		\hline
		$TI1$&Ellrod Indices&$\rm s^{-2}$ \\
		\hline
		$|v|$&Wind Speed &$\rm ms^{-1}$ \\
		\hline
		$|\bigtriangledown_{H}T|$&Horizontal Temperature Gradient& $\rm Km^{-1}$\\
		\hline
		$|v|DEF$&MOS CAT Probability Predictor & $\rm ms^{-2}$\\
		\hline
	\end{tabular}
\label{feature}
\end{table}
\subsection{Features}
In each grid of a feature cube (i.e., $\bsy{X_i}$), we fill it with 12 relevant features (thus $C$=12) as shown in Table \ref{feature}.
The first 6 of them are \ti{raw weather features} while the rest 6 are \ti{turbulence indexes} invented by meteorologists.
Raw features such as temperature, wind component, and pressure can be considered as fundamental features and certain combinations of these features in adjacent areas may contribute to the occurrence of turbulence.
\ti{Deep neural network} such as \ti{convolutional neural network} is capable of learning such complex spatial correlations and it is essential to keep the raw features. 
We further apply 6 turbulence indexes as extra features to enhance the model capacity. 
Most of these indexes are proposed by previous meteorologists,  usually adopted independently or integrated by a weighted sum \cite{sharman2006integrated} in existing turbulence forecasting systems. 
We regard them as prior knowledge and concatenate with raw features. 
\subsection{Labels and the scarcity issue}
We collect turbulence events data from online pilot reports, each is labeled with a severity level. There are four levels in our data, i.e., Negative, Light, Moderate, and Severe.
After gathering the feature data and label data, we align them by time and space.
Details are provided in Section 4.
According to our statistics, at each hour, there are only ${0.05\%}$ grids of North American air space are labeled with a turbulence level while $ {99.95\%}$ are unknown.
Consequently, we have to mask these unlabeled grids during training to bypass the backpropagation of their gradients. 
This leads to less training signals available, making it hard for the network to be trained sufficiently. 
\begin{figure}[t]
\setlength{\belowcaptionskip}{-7pt}
\centering
\includegraphics[width=0.48\textwidth]{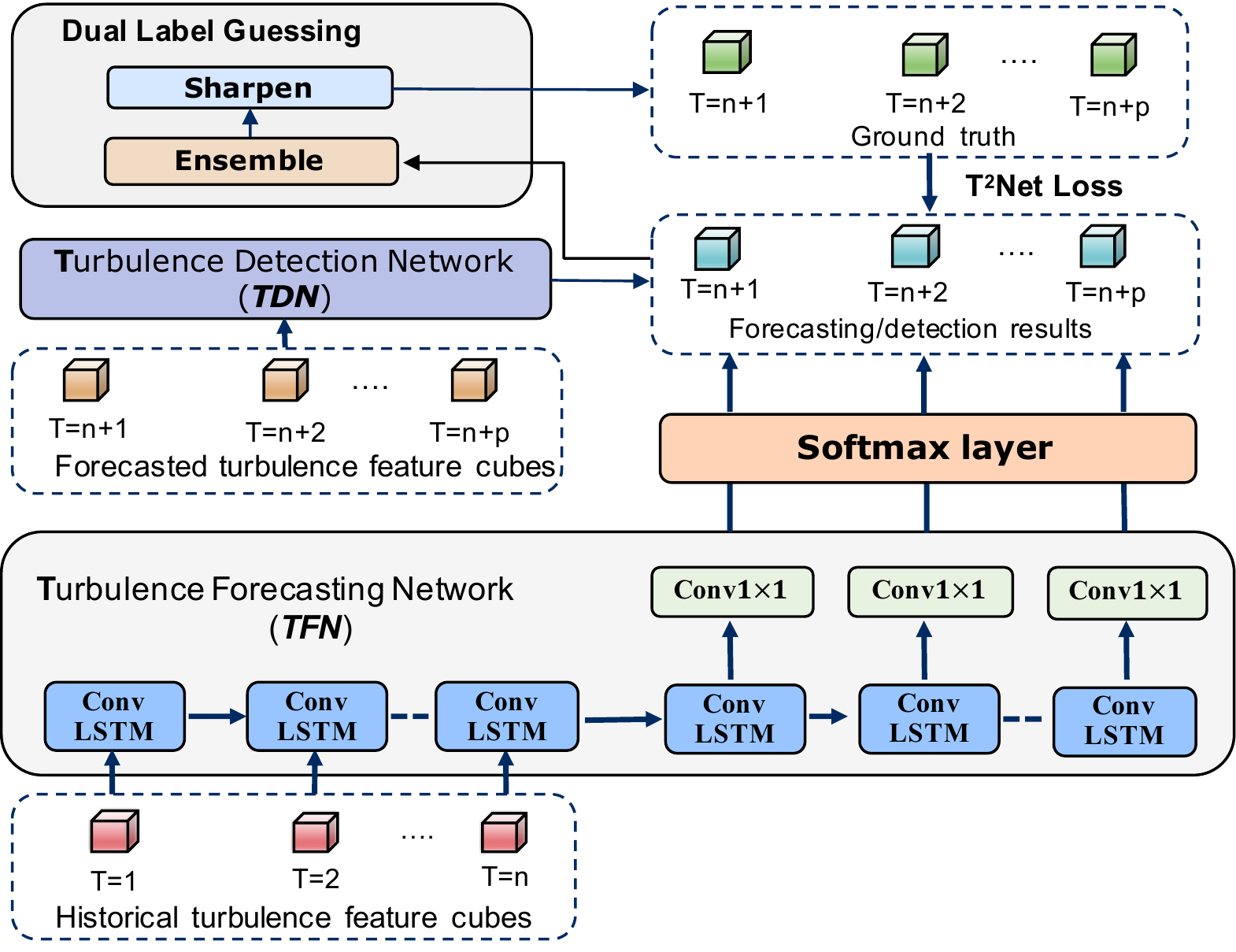}
\caption{The architecture of \model}
\label{framework}
\vspace{-0.1cm}
\end{figure}

\section{Methodology}
In this section, we introduce the details of our proposed turbulence forecasting framework, \model. As shown in Figure \ref{framework}, \model{} mainly consists of a \ti{{\tb{T}}urbulence Forecasting Network} (\tb{T}FN) and a \ti{{\tb{T}}urbulence Detection Network} (\tb{T}DN). 
TFN serves for the main task, i.e., forecasting task, while
TDN serves for the auxiliary task, i.e., turbulence detection.
Based on the predictions of TFN and TDN, a novel \ti{\tb{Dual Label Guessing}} approach is proposed to generate more robust pseudo-labels as additional training data. 
\subsection{Turbulence Forecasting Network}
TFN is designed on top of the basic ConvLSTM \cite{xingjian2015convolutional} architecture to model the complex spatio-temporal correlations among different spatial grids.
ConvLSTM is a variation of LSTM which extends basic LSTM cell by replacing the fully connected layer with convolution operation in the internal transitions.
As shown in Figure \ref{framework}, TFN consists of two ConvLSTMs, serving as the encoder and decoder respectively. 
The encoder takes a sequence of 4D tensors as input, $\bsy{X_1,X_2,...,X_n} \in \mathbb{R}^{\mc{L\times W\times H \times C}}$, i.e., the historical turbulence feature cubes of time slots $1,...,n$. 
The decoder takes the last hidden state of the encoder as the initial hidden state, and uses \ti{teacher forcing} \cite{goodfellow2016deep} (use previous ground truth $\bsy{Y_{j-1}}$ as the next input to the decoder) to generate a sequence of features corresponding to the forecasting time slots $n+1,...,n+p$. 
The decoder's outputs are then fed into to a $\text{Conv}1\times 1$ block followed with a Softmax layer to produce the forecasted turbulence levels $\bsy{P_{n+1},P_{n+2},...,P_{n+p}} \in \mathbb{R}^{\mc{L\times W\times H\times} 4}$. 
The process of TFN can be summarized as:
\begin{equation*}
\bsy{h_i^{enc},o_i^{enc}} = \text{ConvLSTM}^{enc}\big(\bsy{X_{i},h_{i-1}^{enc}}\big), i\in[1,n]
\end{equation*}
\begin{equation*}
\bsy{h_j^{dec},o_j^{dec}} = \text{ConvLSTM}^{dec}\big(\bsy{Y_{j-1}, h_{j-1}^{dec}}\big),\text{$j\in$[$n$+1,$n$+$p$]}
\end{equation*}
\begin{equation*}
\bsy{P_j} = \text{Softmax}\big(\text{Conv1}\times 1(\bsy{o_{j}^{dec}})\big),\text{$j\in$[$n$+1,$n$+$p$]}
\end{equation*}
\subsection{Turbulence Detection Network}
The Turbulence Detection Network (TDN) employs \ti{Convolutional Neural Network} to extract spatial correlations and detect the turbulence levels. The input to TDN is the NWP forecasted turbulence feature cube $X_i$ at time slot $i$, and the output is the detected turbulence level cube $P_i \in \mathbb{R}^{\mc{L\times W\times H } \times 4}$ at the same time slot. 
TDN can be summarized as:
\begin{equation}
\text{Conv}\big(\bsy{X_{i}},1\big) = f_1\big(\bsy{X_i}\circledast \bsy{W_1}+b_1\big),
\end{equation}
\begin{equation}
\text{Conv}\big(\bsy{X_{i}},l\big) = f_l\big(\text{Conv}(\bsy{X_{i}},l-1)\circledast \bsy{W_l}+b_l\big),
\end{equation}
\begin{equation}
\bsy{P_i} = \text{Softmax}\big(\text{Conv}(\bsy{X_{i}},l)\big),\text{$i\in$[$n$+1,$n$+$p$]},
\end{equation}
where $l$ denotes the l-th layer, $f_l$ denotes the activation function of l-th layer, ``$\circledast $'' denotes the {3D-convolution} operator.

\subsection{Dual Label Guessing}
\begin{figure}[t]
\setlength{\belowcaptionskip}{-6pt}
\centering
\includegraphics[width=0.48\textwidth]{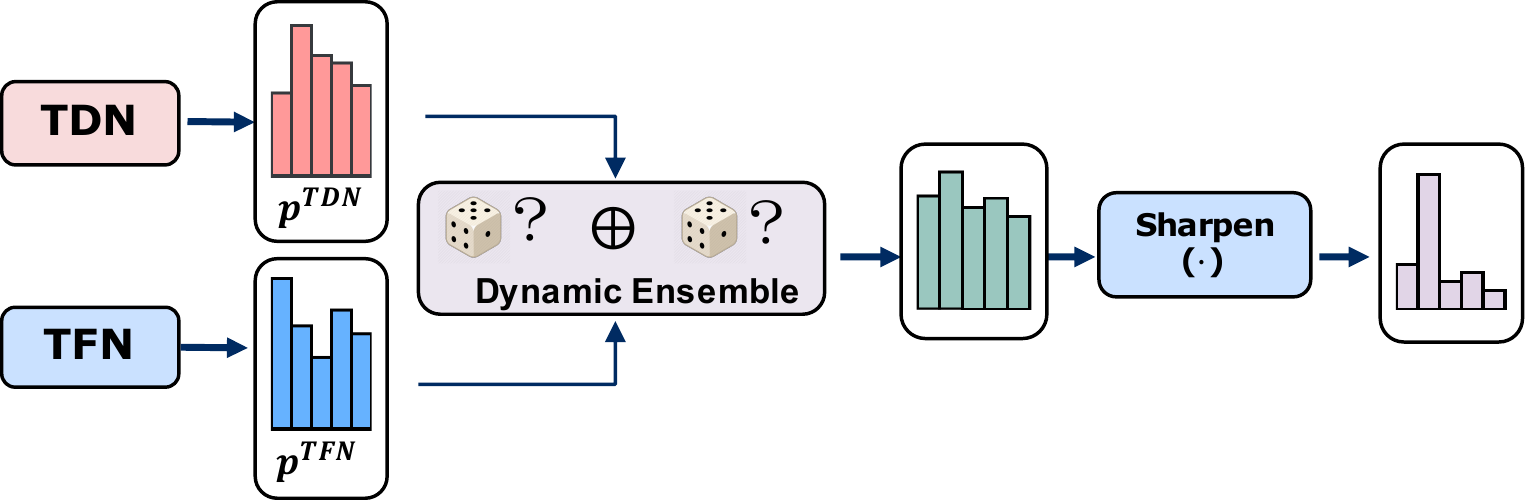}
\caption{Diagram of dual label guessing}
\label{dual}
\vspace{-0.1cm}
\end{figure}

To mitigate the label scarcity issue, we propose \ti{Dual Label Guessing} (DLG), as illustrated in Figure \ref{dual}. 
During training, DLG will generate pseudo-labels for those unlabeled grids so that we can obtain additional training data. 
To highlight, DLG differs from existing ``label-guessing'' methods \cite{lee2013pseudo, berthelot2019mixmatch} in two ways:
\begin{itemize}[leftmargin=*]
\item \ti{\tb{Complementary Dual Semi-supervised Signals.}}
Instead of \ti{single source} inference, our method leverages \ti{dual source} signals from two related but different tasks.
DLG combines the predictions from TDN and TFN, protecting each other from their individual errors/bias, thus getting more robust to generate high-quality pseudo-labels.   

\item \ti{\tb{Soft Labels.}} Instead of the \ti{hard label} in other approaches like ``pseudo-labeling'' \cite{lee2013pseudo} which takes the class with the highest probability and produce a one-hot label, we produce \ti{soft label} via a ``sharpening'' function \cite{berthelot2019mixmatch}, yielding a class distribution. 
The soft label is smoother and more error-tolerant compared with hard label. 



\end{itemize}

\subsubsection{\tb{{{Dynamic Ensemble of TDN and TFN}}}}
In our {Dual Label Guessing}, we first propose a novel \ti{Dynamic Ensemble} method to fuse the predictions of TFN and TDN grid by grid, the combined prediction is defined as: 
\begin{equation}
 \bsy{p}=\frac{{\bs }(\bsy{p^{\sss TDN},p^{\sss TFN}},\tau(t))\oplus {\bs }(\bsy{p^{\sss TDN},p^{\sss TFN}},\tau(t))}{2}
\end{equation}
where $\bsy{p^{\sss TDN}, p^{\sss TFN}} \in \mathbb{R}^{4}$ are output vectors of a single grid predicted by TDN and TFN respectively, in which each element represents the probability of each turbulence class. $\oplus$ denotes element-wise addition.
$\bs$ denotes the $\ti{\tb{binary sampling}}$. 
To be noted, two ${\bs }(\bsy{p^{\sss TDN},p^{\sss TFN}},\tau(t))$ in the equation are different samples and the sampling function $\bs$ is defined as:
\begin{equation}
\bs(\bsy{p^{\sss TDN}, p^{\sss TFN}},\tau(t)) = \left\{\begin{matrix}
\bsy{p^{\sss TDN}}, & \text{if } r(t) >\tau(t)\\ 
\bsy{p^{\sss TFN}}, & \text{if } r(t)<=\tau(t) 
\end{matrix}\right.
\end{equation}
$r(t)$ above is a \ti{pseudorandom number} between $[0,1]$ with $t$ as the seed. 
$\tau(t)$ is a \ti{dynamic coefficient} controlling the probability of drawing $\bsy{p^{\sss TDN}}$ or $\bsy{p^{\sss TFN}}$ , i.e., \ti{relative importance} of TDN and TFN, $\tau(t)$ is defined as a piece-wise function:
\begin{equation}
\tau(t) = \left\{\begin{matrix}
 0 & t<T_1\\ 
 \frac{t-T_1}{T_2-T_1}\beta & T_1<t<T_2 \\ 
 \beta & t>T_2 
\end{matrix}\right.
\end{equation}

\noindent where $t$ is the number of epochs, $T_1,T_2$ and $\beta$ are hyperparameters. 
The design of $\tau(t)$ follows the intuitions:
at the beginning of training, TDN shall have a higher probability  (in the first stage, $1-\tau(0)=1$ makes TDN 100\% to be chosen), because TDN is pre-trained, predicting more accurately than TFN. 
As the iteration $t$ increases gradually, TDN's probability should decrease and TFN's increases since TFN's accuracy is growing. 
Finally, the binary sampling probability stabilizes at some balancing point $\beta\in (0,1]$.


\subsubsection{\tb{Soft Labels}}
After getting the ensembled prediction $p$, to obtain the pseudo-label, we further apply a sharpening function to minimize the \ti{entropy} of the label distribution, which is defined as: 
\begin{equation}
\text{Sharpen}(\bsy{p},T)_i := \bsy{p}[i]^{\frac{1}{\sss T}}\Big/\sum_{j=1}^{4}\bsy{p}[j]^{\frac{1}{\sss T}}
\end{equation}
where $\bsy{p}[i]$ is the $i$-th element of $\bsy{p}$, $T$ is a hyper-parameter to adjust the ``temperature'' of this categorical distribution. $\tb{Sharpen}(\bsy{p},T)$ first calculates the $T$-th power of each elements and then based on which performs a normalization. When $T\rightarrow 0$, the result will approach a one-hot distribution.

\subsection{Loss Function}
The loss function of our \model{} includes two parts: (1) $\mc{L}_s$, the supervised part for the labeled grids, (2) $\mc{L}_u$, the unsupervised part for the grids with pseudo-labels. 
\begin{equation}
\mc{L}=\mc{L}_s+\lambda\mc{L}_u
\end{equation}
where $\lambda\in[0,1]$ is a hyperparameter controlling the weight of unsupervised loss. For $\mc{L}_s$, we adopt cross-entropy, and for $\mc{L}_u$ , we employ the L2 distance between model predictions and pseudo labels.

\section{Experiments}
\newcommand{\tabincell}[2]{\begin{tabular}{@{}#1@{}}#2\end{tabular}}
\begin{figure*}
\centering
\subfigure[Accuracy]{\label{a1}\includegraphics[trim=0 20 0 0, clip, width=3.9cm]{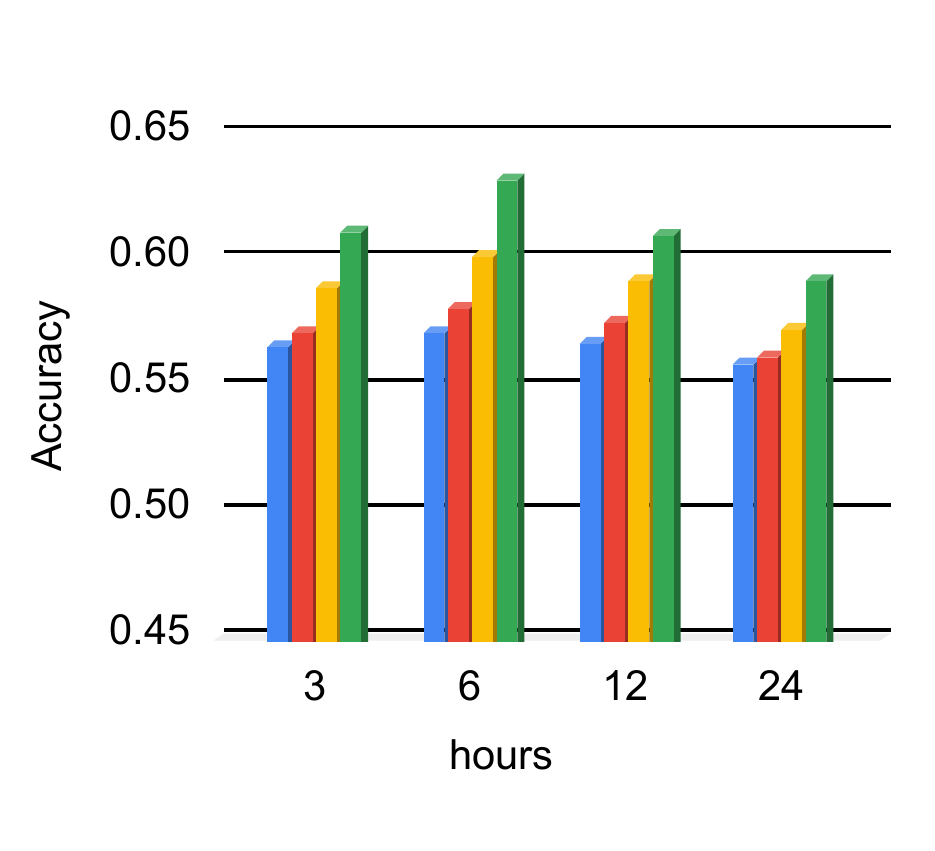}}
\subfigure[Weighted-Precision]{\label{a2}\includegraphics[trim=0 20 0 0, clip, width=3.9cm]{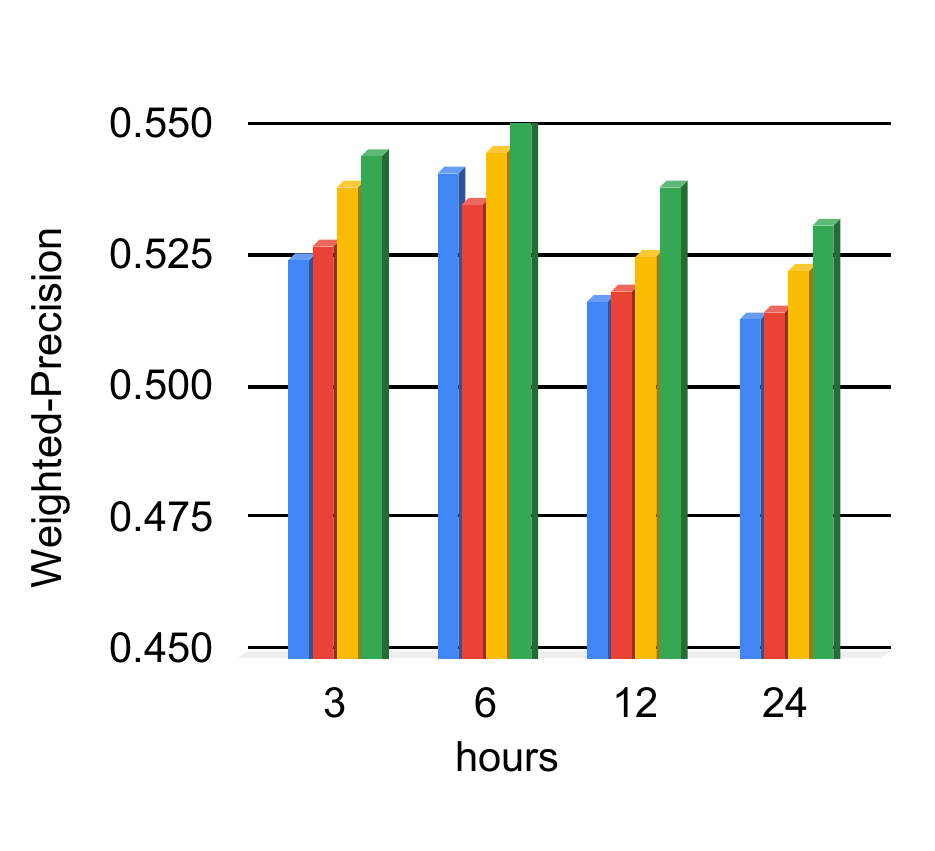}}
\subfigure[Weighted-Recall]{\label{a2}\includegraphics[trim=0 20 0 0, clip, width=3.9cm]{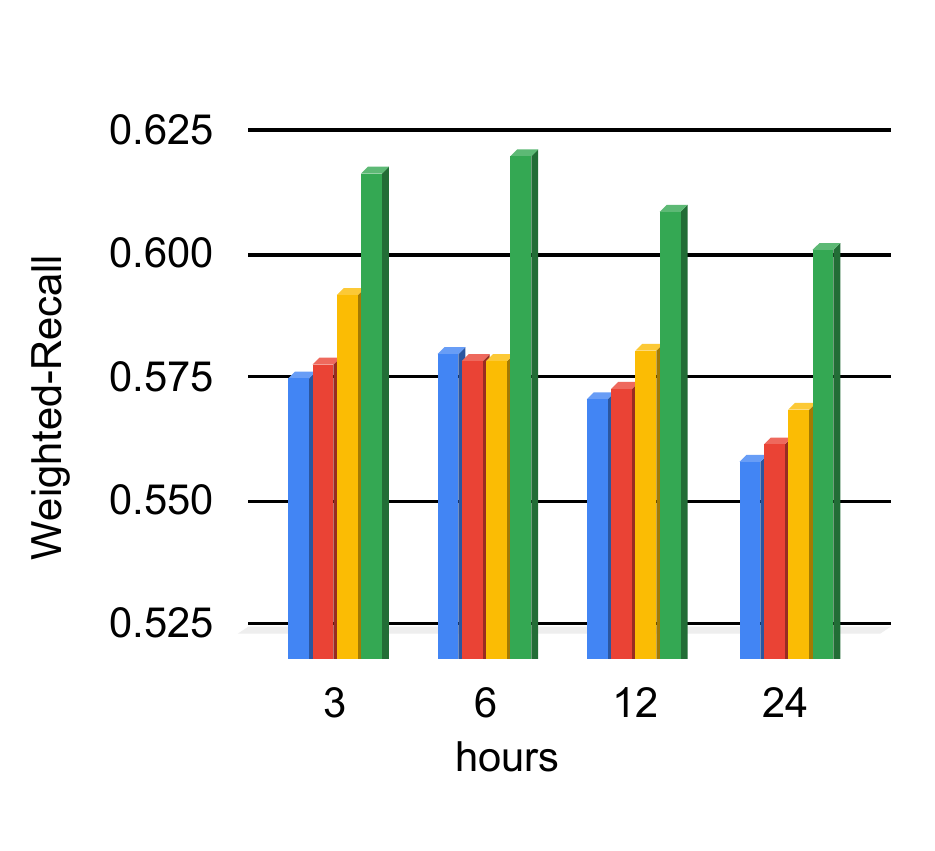}} 
\subfigure[Weighted-F1]{\label{a2}\includegraphics[trim=0 20 0 0, clip, width=4.7cm]{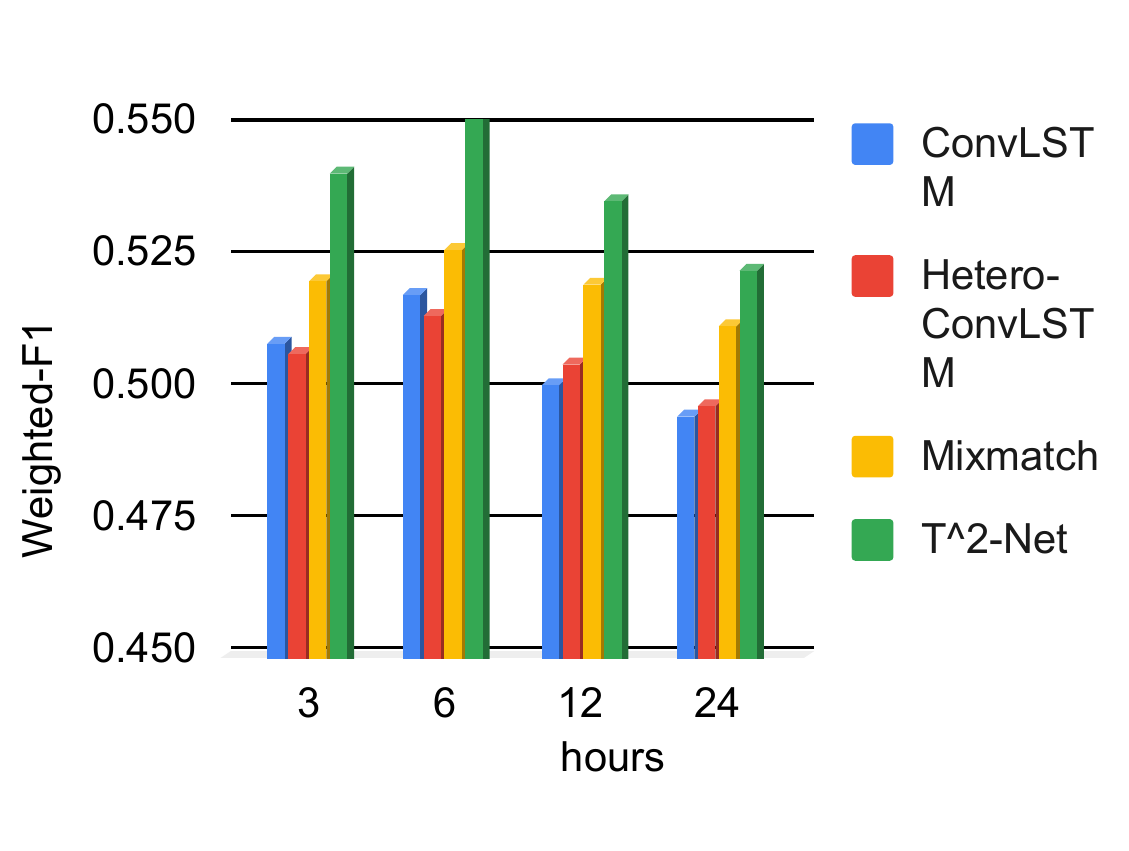}}  
\caption{Performance comparison of different forecasting lengths.}\label{hours}
\vspace{-6mm}
\end{figure*}
In this section, we first introduce the data and settings of our experiments. 
Then we evaluate the performances of the proposed forecasting model on different forecasting lengths compared with several state-of-the-art baselines. 
Lastly, we show the parameter sensitivity analysis.
\subsection{Experimental Settings}
\noindent\tb{Data Preprocessing:} All the data we use in the experiment are publicly available. 
We first obtain
 30 days of weather data (from 20190601 to 20190630) 
from National Oceanic and Atmospheric Administration (NOAA) \footnote{ftp://nomads.ncdc.noaa.gov/RUC/13km/201906/}, then obtain 30 days of turbulence report data (the same period as the weather data) from Iowa Environmental Mesonet (IEM) \footnote{http://mesonet.agron.iastate.edu/request/gis/pireps.php}. 
The weather data is generated hourly on a 13-km (8-mile) resolution horizontal grid, with $451\times337$ grids in total, representing the North American region. 
In vertical direction, there are 36 different geopotential heights.
We treat the whole data as a \ti{$451\times337\times36$} cube and use a sliding window of size $10\times10\times5$ to generate the hourly \ti{feature cubes} (each grid contains the raw weather features and the turbulence indexes). 
For the turbulence report, each report contains the coordinate, geopotential height, time, and level of the turbulence, and we use the same way to generate the hourly \ti{label cubes} (each grid contains the turbulence level). 
We filter out the cubes NOT in the ``cruising altitude'' (31000 to 38000 feet) since most pilot reports are recorded among these heights. 
Finally, we use a sliding window of $n+p$ hours, adopt the first $n$ hours' feature cubes as the input sequence and the next $p$ hours' label cubes as the output sequence to generate the training data. 
In experiments, we investigated different $n$ and $p$, specifically, $n=p=\{3,6,12,24\}$.
We further randomly split the data into training/validation/testing set with the ratio of {6:2:2}.


\begin{table}
\small
\caption{Overall performance of $\text{T}^{2}$-Net and baselines.}
	\centering
	\begin{tabular}{c|c|c|c|c}
		\hline
		\hline
		\multirow{2}*{Method} & \multirow{2}*{Accuracy} & \multicolumn{3}{c}{Weighted}   \\
		\cline{3-5}
		~ & ~  & P &R & F1 \\
		\hline
		\hline 
		 TBI (Turbulence indexes)  &0.428  &0.368  &0.428 &0.382  \\ 
		 Multinomial LR  &0.434  &0.355  &0.434 &0.385  \\ 
		 MLP (3-layers) &0.449  &0.393  &0.449 &0.355  \\  
		 GBDT (100 trees) &0.440  &0.370 &0.440 &0.341 \\  
		  Attentional LSTM &0.489 & 0.378 &0.489 & 0.426 \\ 
		  CNN (kernel size=3) &0.491 & 0.437 &0.491 & 0.370 \\  
		 ConvLSTM &0.571   &0.548  &0.581 &0.518  \\ 
		 Hetero-ConvLSTM &0.580   &0.536  &0.580 &0.520  \\ 
		\hline
		 Pseudo-labeling &0.591   &0.536  &0.600 &0.536  \\
		 Mixmatch &0.600   &0.546  &0.580 &0.540  \\
		\hline
		 $\text{T}^{2}$-Net &\tb{0.623} & \tb{0.551} &\tb{0.614}  &\tb{0.548} \\
		\hline
		\hline
	\end{tabular}
\label{overall}
\vspace{-6mm}
\end{table}

\noindent\tb{Evaluation Metrics: }
We adopt several standard multi-class classification metrics to evaluate all the models: (1) Accuracy, (2) Weighted-Precision, (3) Weighted-Recall, (4) Weighted-F1. 
We calculate these metrics by averaging them throughout all the labeled grids in different time steps and skip the grids labeled with ``unknown''.

\noindent\tb{Parameter Configuration: }For our model and all the baseline methods, we obtain the optimal parameters on the validation set using early stopping. 
For TFN, we adopt 1-layer ConvLSTM with $3\times3\times3$ kernel for both encoder and decoder, using Sigmoid as activation function. For TDN, we adopt 3-layer CNN with  $3\times3\times3$, $3\times3\times3$ and $5\times5\times3$ kernels respectively, using Relu as inner activation function.
The optimal hyperparameters of the rest part are $\beta=0.6, T_1=5, T_2=15, T=0.5, \lambda=0.4$.

\noindent\tb{Baseline Methods: }To systematically investigate the performance of modern machine learning methods on turbulence forecasting, we compare our proposed \model{} with 3 categories of baselines. (1) \ti{Tubulence indexes} (\tb{TBI}) \cite{sharman2006integrated}, an integrated approach which combines multiple turbulence indexes to forecast turbulence. (2) \ti{Supervised learning methods}: a number of supervised machine learning methods are examined: Multinomial Logistic Regression (\tb{Multinomial LR}), Multi-layer Perceptron (\tb{MLP}), Gradient Boosting Decision Tree (\tb{GBDT}) \cite{friedman2001greedy}, \tb{Attentional LSTM}, Convolutional Neural Network (\tb{CNN}), \tb{ConvLSTM} \cite{xingjian2015convolutional} and \tb{Hetero-ConvLSTM} \cite{yuan2018hetero}. 
We test all these methods using the same base features as our model, i.e., the 6 raw weather features and 6 turbulence indexes. (3) \ti{Semi-supervised learning (SSL) methods}: we also compare with several state-of-the-art semi-supervised learning methods since abundant unlabeled data exists in the turbulence forecasting task and \model{} is also semi-supervised. \tb{Pseudo-labeling} \cite{lee2013pseudo}, a simple yet effective SSL method which retrains the model with pseudo-labels predicted by the model itself. 
\tb{Mixmatch} \cite{berthelot2019mixmatch}, a recent holistic approach which unifies several dominant SSL methods and achieves state-of-the-art results on 4 benchmark datasets. To ensure a fair comparison, we apply the same base model (ConvLSTM) for these SSL methods.


\begin{figure*}[htbp]
\centering
\subfigure[$\beta$]{\label{a1}\includegraphics[ width=4.8cm]{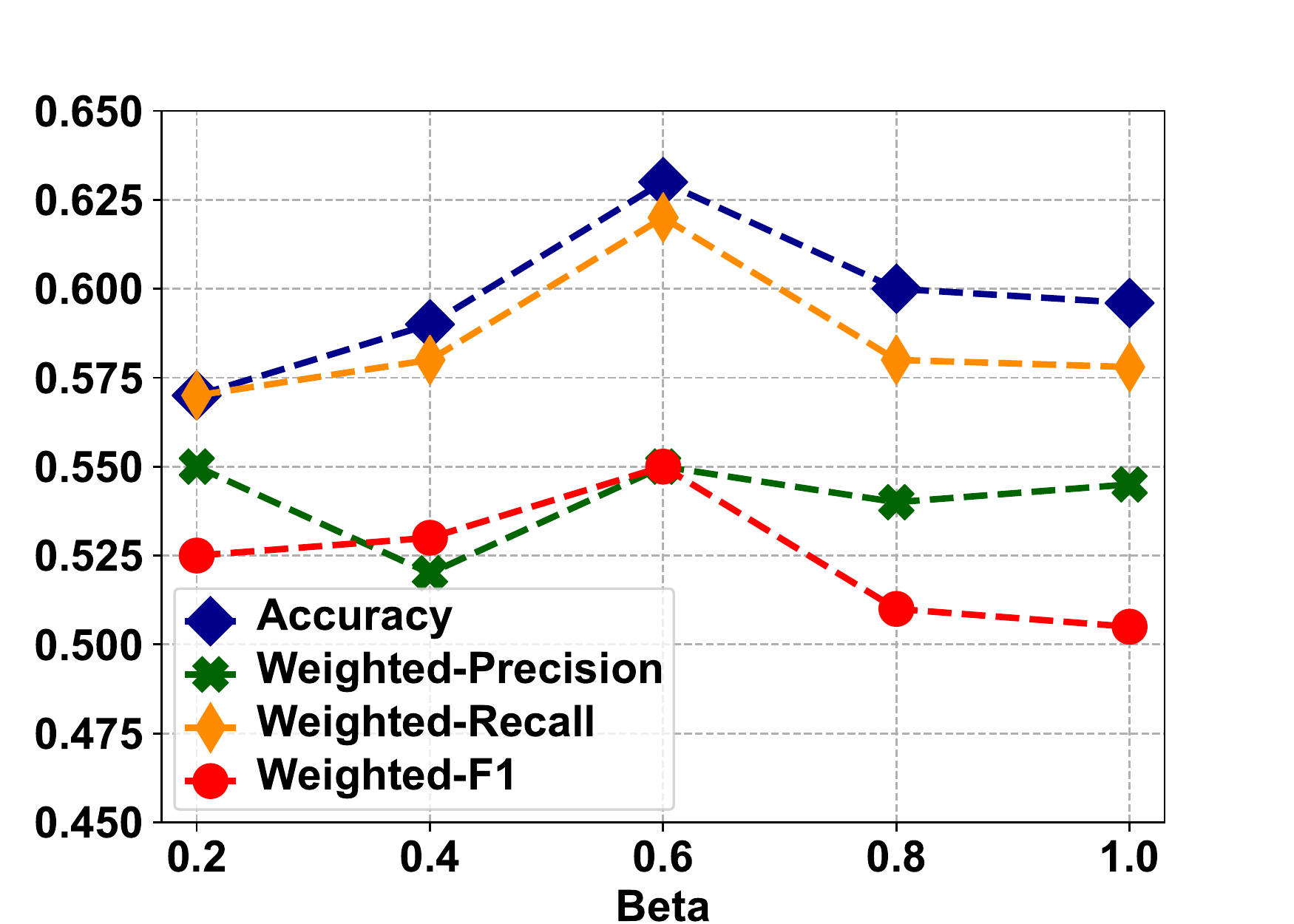}}
\subfigure[T]{\label{a2}\includegraphics[ width=4.8cm]{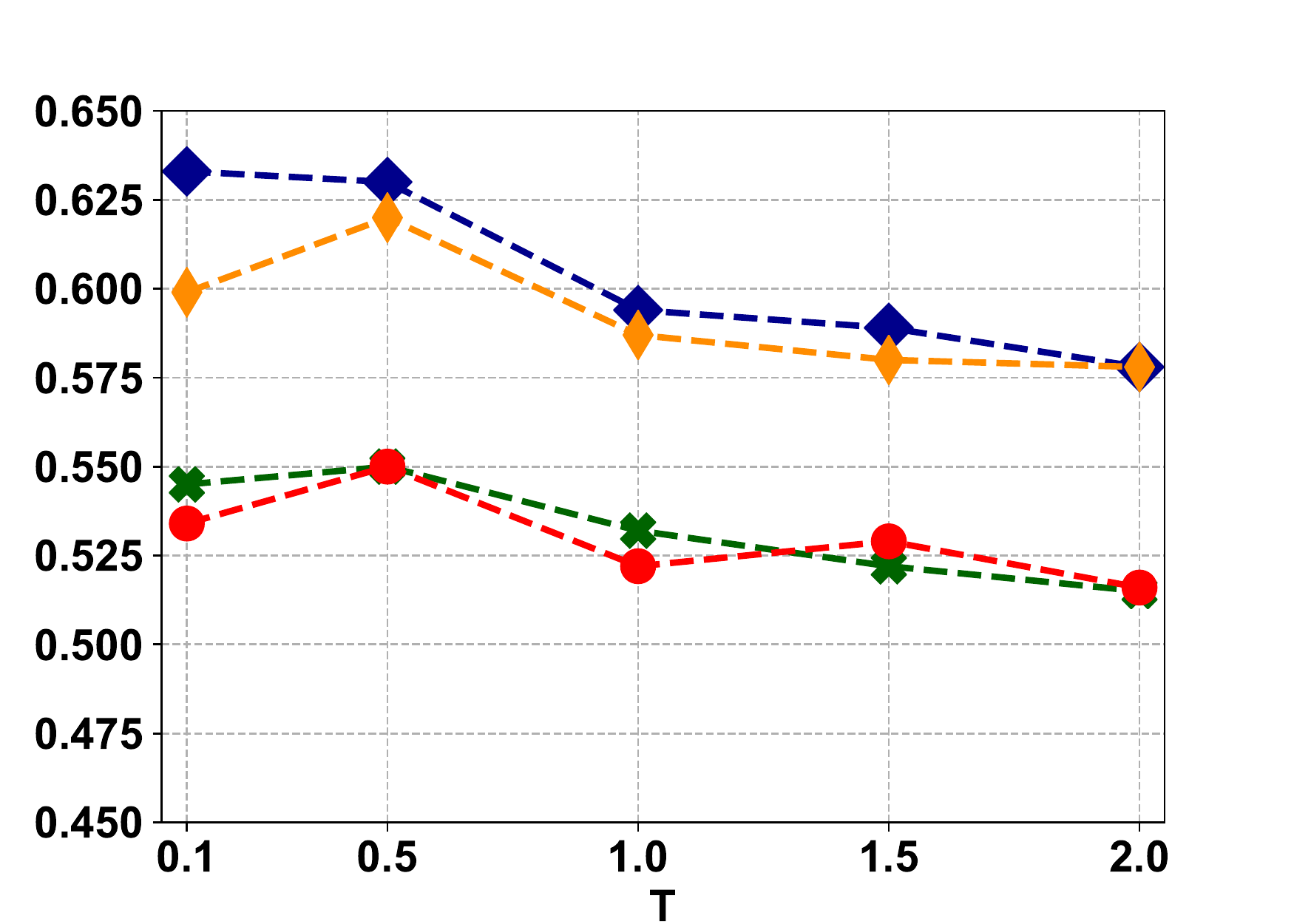}}
\subfigure[$\lambda$]{\label{a2}\includegraphics[ width=4.8cm]{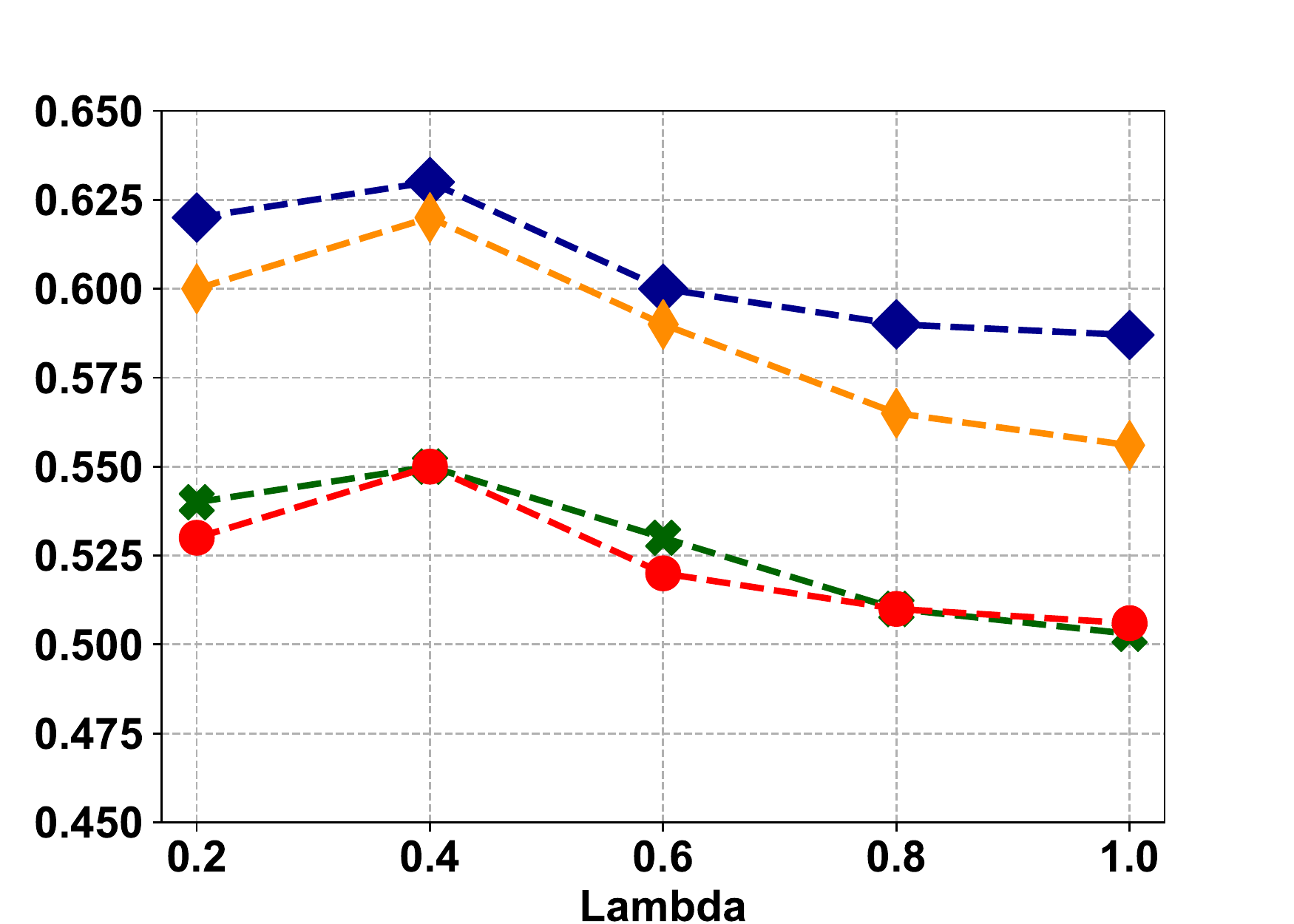}} 
\caption{Parameter sensitivity analysis.}\label{parameter}
\vspace{-6mm}
\end{figure*}
\subsection{Overall Results on Turbulence Forecasting}
We first present the overall performance of all baselines and \model{} when $n=p=6$. 
The results are presented in Table \ref{overall}. 
MLP and GBDT perform better than TBI and Multinomial LR because they integrate nonlinearity. 
By taking temporal correlation and spatial correlation into consideration, Attentional LSTM and CNN achieve better accuracy than those not. 
ConvLSTM further improves the performance because of modeling spatial and temporal correlation simultaneously. 
Hetero-ConvLSTM achieves the best accuracy and Weighted F1 among supervised methods for it predicts based on an ensemble of multiple ConvLSTMs of different geographical areas. 
However, Hetero-ConvLSTM has an efficiency issue for training many ConvLSTMs at the same time.
For the semi-supervised learning baselines, all of them beat the supervised methods, verifying the practicability of taking advantage of unlabeled data. 
Among them, Mixmatch achieves the best accuracy, Weighted F1, Weighted Precision, and thus proves that \ti{soft pseudo-label} is better than \ti{hard pseudo-label}.  
\model{} achieves the best performance with the highest Accuracy, Weighted Precision/Recall/F1. 
This indicates that on turbulence forecasting task, \model{} has superior ability to model the complex spatio-temporal relation as well as utilizing the abundant unlabeled data to enhance training.

We also report the overall performance of \model{} and several representative baselines (ConvLSTM, Hetero-ConvLSTM, and Mixmatch) for different forecasting lengths ($n=p=3,6,12,24$). 
As shown in Figure \ref{hours}, we can observe that \model{} achieves the best performance on different time lengths. 
However, an interesting phenomenon brought to our attention is that when $n,p$ increases ($>$6), there is a drop in performance for all the baseline models and \model. 
We attribute this to the increased complexity and ``gradient vanishing'' problem in long-sequence prediction. 
However, performance on $n,p=6$ is better than $n,p=3$, this is because the model benefits more from the increased input feature length when the time length changes from 3 to 6.

\subsection{Parameter Sensitivity Analysis}
Figure \ref{parameter} presents the sensitivity analysis of the key parameters of our Dual Label Guessing
and loss function, i.e., $\beta, T,$ and $\lambda$. 
We obtained the sensitivity curve of each parameter by fixing the rest using their optimal values. 
We can observe that the best performance is achieved when $\beta=0.6$ , this indicates TFN is more important to the final result. 
Besides, $\lambda=0.4$ achieves the best results indicating that there is a trade-off of utilizing the unlabeled data.
We can also observe that the performance is relatively stable as the parameters change, thus proves the robustness of \model.

\section{Related Work}
Traditional turbulence forecasting approaches mainly focus on devising various turbulence indexes \cite{colson1965index,dutton1980probability,ellrod1992objective}.
Sharman et al. reviewed 13 turbulence indexes and proposed an integrated method combining these features to achieve further improvements \cite{sharman2006integrated}.
However, as validated in our experiments, methods solely rely on turbulence indexes can not achieve satisfactory performance. 
Our method takes advantage of both turbulence indexes and the power of deep learning to achieve superior performance.
Recent years have witnessed a growing interest in applying machine learning to a variety of spatio-temporal problems \cite{karpatne2016monitoring,yuan2020spatio,xingjian2015convolutional}.
Most of them adopt ConvLSTM to model the spatio-temporal correlations without facing the label scarcity issue.
Particularly, we propose a novel semi-supervised approach to tackle the scarcity issue considering by taking complementary turbulence signals into account.
More recently, some work tries to incorporate physical principles with deep learning to facilitate turbulent flow modeling, however, they are not directly tackling turbulence forecasting, but simulating and predicting certain turbulence variables, e.g., Wang et al. predicts the velocity fields \cite{wang2019towards}.
To our best knowledge, our work is the first systematic machine learning study, that directly forecasts the occurrence of flight turbulence, using sparse turbulence labels extracted from pilot reports as supervision.

\section{Conclusion}
In this paper, we developed a data-driven framework for turbulence forecasting. Specifically, we first built an encoder-decoder paradigm based on ConvLSTM to model the spatio-temporal correlations. Then, to address the label scarcity issue, we proposed a novel {Dual Label Guessing} method, which integrated complementary signals from the main task of {\tb{T}}urbulence Forecasting and the auxiliary task of {\tb{T}}urbulence Detection to generate pseudo-labels. 
Finally, we conducted extensive experiments on a real-world dataset which showed that the proposed approach can greatly alleviate the problem of spatio-temporal correlation modeling as well as label scarcity on turbulence forecasting.


\bibliographystyle{ieeetr}
\bibliography{ref}

\vspace{12pt}

\end{document}